\documentclass{article} % For LaTeX2e
\usepackage[final]{colm2025_conference}

\usepackage{microtype}
\usepackage{hyperref}
\usepackage{url}
\usepackage{booktabs}
\usepackage{amsmath}
\usepackage{lineno}
\usepackage{graphicx}
\usepackage{wrapfig}

\definecolor{darkblue}{rgb}{0, 0, 0.5}
\hypersetup{colorlinks=true, citecolor=darkblue, linkcolor=darkblue, urlcolor=darkblue}

\definecolor{customcolor}{rgb}{0.675, 0.399, 0.373}
\hypersetup{
    colorlinks=true, % Enables colored links
    linkcolor=blue, % Internal links
    citecolor=blue  % Citation links
}

\usepackage{tcolorbox}
\tcbuselibrary{listings,breakable,xparse,skins}  % Use listings instead of minted

\definecolor{bg}{HTML}{1F1D2E}
\definecolor{codetype}{HTML}{89DCEB}    % For Pipeline, Layer, etc.
\definecolor{codestring}{HTML}{A6E3A1}  % For strings like "gpt-4o"
\definecolor{linenumbers}{HTML}{9792A2}

\usepackage{fvextra}
\usepackage{inconsolata}  % Alternative monospaced font for pdfLaTeX

\definecolor{codebg}{rgb}{0.12, 0.12, 0.15}  % Dark background
\definecolor{codetext}{rgb}{0.85, 0.85, 0.85}  % Light gray text
\definecolor{keywordcolor}{rgb}{0.98, 0.50, 0.30}  % Bright orange for keywords
\definecolor{stringcolor}{rgb}{0.60, 0.85, 0.50}  % Green for strings
\definecolor{commentcolor}{rgb}{0.55, 0.55, 0.60}  % Dim gray for comments
\definecolor{numbercolor}{rgb}{0.90, 0.80, 0.40}  % Yellow for numbers
\definecolor{operatorcolor}{rgb}{0.85, 0.60, 0.90}  % Purple for operators
\definecolor{triplequote}{rgb}{0.60, 0.85, 0.50}  % Green for triple-quoted strings
\definecolor{funcname}{rgb}{0.85, 0.60, 0.90}  % Purple for function and class names

\lstdefinestyle{pythonstyle}{
    language=Python,
    backgroundcolor=\color{codebg},
    basicstyle=\fontsize{8}{8}\ttfamily\color{codetext},  % Main text color
    keywordstyle=\bfseries\color{keywordcolor},  % Built-in Python keywords (bold + orange)
    stringstyle=\color{stringcolor},  % Strings
    commentstyle=\itshape\color{commentcolor},  % Comments
    numberstyle=\color{numbercolor},  % Line numbers
    emphstyle=\color{funcname},  % Functions and class names (purple, NOT bold)
    emph={DiscreteScale, Pipeline, Layer, CoTUnit, JudgeUnit, WeightedSummedScoreExtractor, MeanVariancePoolUnit, MeanPoolUnit, ConversationalUnit, BooleanScale, prompt, extract, via, pin},  % Functions & classes in purple
    emphstyle={[2]\color{keywordcolor}},  % Custom keywords (orange, NOT bold)
    emph={[2]temperature, retries, repeat, role_name, inner, outer},  % Custom keywords
    literate={"""}
        {{\textcolor{triplequote}{\texttt{"""}}}}3  % Triple-quoted strings highlighted
        {'}{\textquotesingle}1
        {""}{\textquotedbl}1
        {>=}{{\textcolor{operatorcolor}{\texttt{>=}}}}2
        {<=}{{\textcolor{operatorcolor}{\texttt{<=}}}}2
        {==}{{\textcolor{operatorcolor}{\texttt{==}}}}2
        {=>}{{\textcolor{operatorcolor}{\texttt{=>}}}}2,
    breaklines=true,
    frame=none,
    showstringspaces=false,
    tabsize=4,
    upquote=true,
    columns=flexible,
    keepspaces=true
}

\NewTCBListing{mintedbox}{O{}m!O{}}{%
    reset,  
    breakable=true,
    listing engine=listings,
    listing only,
    boxsep=0pt,
    left=10pt,
    right=10pt,
    top=5pt,
    bottom=5pt,
    arc=4pt,
    colback=codebg,
    colframe=codebg,
    enhanced,
    overlay={%
        \begin{tcbclipinterior}
            \fill[codebg] (frame.south west) rectangle (frame.north east);
        \end{tcbclipinterior}
    },
    listing options={%
        style=pythonstyle,  % Use our defined Python style
        #1
    }
}

\newcounter{snippetpython}

\makeatletter
\lstnewenvironment{snippetpython}[1][]{%
    
    \let\c@lstlisting=\c@snippetpython  %
    
    \lstset{
        language=Python,
        basicstyle=\ttfamily\footnotesize,
        captionpos=b,                    %
        showstringspaces=false,          %
        breaklines=true,
        numbers=left,
        numberstyle=\tiny\color{gray},
        keywordstyle=\color{DarkCyan},
        stringstyle=\color{OrangeWeb},
        commentstyle=\color{Avocado},
        morekeywords={dspy, self},       %
        backgroundcolor=\color{gray!5},
        numbersep=5pt,                   %
        xleftmargin=20pt,                %
        xrightmargin=20pt,               %
        frame=single,
        framerule=0pt,
        framesep=10pt,                   %
        framexleftmargin=10pt,           %
        framexrightmargin=10pt,          %
        #1
    }
}{}
\makeatother

\newcommand{\textttpop}[1]{\text{#1}} %tt{\textcolor{customcolor}{#1}}}
\newcommand{\textscpop}[1]{\text{#1}} %sc{\textcolor{customcolor}{#1}}}

\title{\textit{Verdict}: A Library for Scaling Judge-Time Compute}

\author{Nimit Kalra, Leonard Tang\\
Haize Labs, NYC\\
\texttt{\{nimit, leonard\}@haizelabs.com} \\
}

\usepackage[T1]{fontenc}
\usepackage{textcomp}
\begin{document}

\ifcolmsubmission
\linenumbers
\fi
\maketitle

\begin{figure}[h!]
    \centering
    \includegraphics[width=\linewidth]{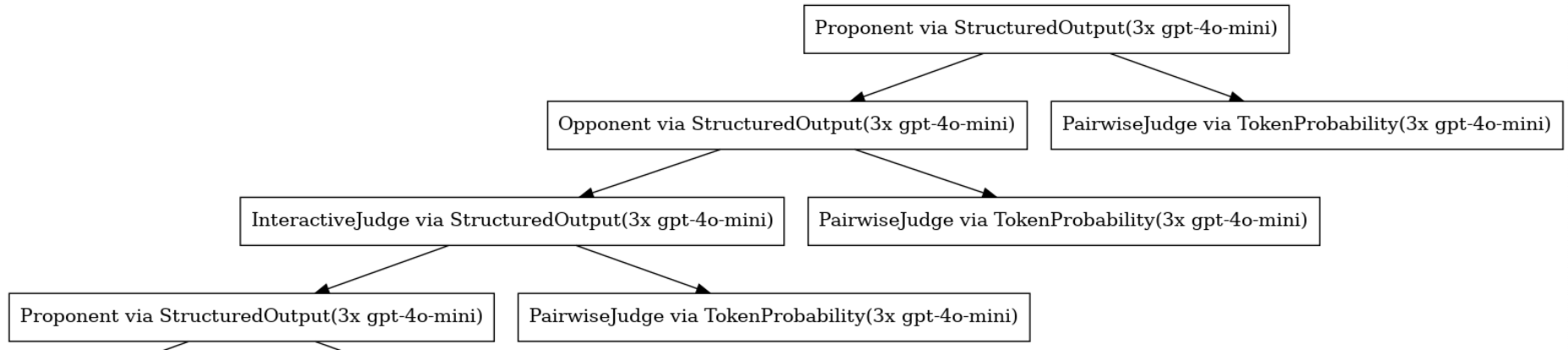}
    \caption{In the Interactive Debate protocol \citep{khan2024debating}, independent instances of a language model adopt opposing positions on an evaluation query (e.g., \textit{Is the following response harmless and helpful?}). After each round, a separate model summarizes the debate. \textbf{Verdict} allows for straightforward implementation of this protocol through its declarative interface for defining and composing modules for parallel execution. This enables flexible scaling of test-time compute at both the component level (e.g., model capacity or role) and the architectural level (e.g., length or trace shape).}
    \label{fig:placeholder}
\end{figure}

\begin{abstract}
The use of LLMs as automated judges ("LLM-as-a-judge") is now widespread, yet standard judges suffer from a multitude of reliability issues. To address these challenges, we introduce \textscpop{Verdict}, an open-source\footnote{Code available at \url{https://github.com/haizelabs/verdict}.} library for scaling \textit{judge-time compute} to enhance the accuracy, reliability, and interpretability of automated evaluators. \textscpop{Verdict} leverages the composition of modular reasoning units---such as verification, debate, and aggregation---and increased inference-time compute to improve LLM judge quality. Across a variety of challenging tasks such as content moderation, fact-checking, and hallucination detection, \textscpop{Verdict} judges achieves performance competitive with orders-of-magnitude larger fine-tuned judges, prompted judges, and reasoning models. Our framework establishes a foundation for scalable, interpretable, and reliable LLM-based evaluation systems for both researchers and practitioners.
\end{abstract}

\section{Introduction}
Automated evaluation using LLMs, a.k.a. ``LLMs-as-a-judge'', is a widely adopted practice for both developers and researchers building LLM-powered applications. However, LLM judges still face a variety of reliability issues, such as inconsistent output formats, missing or miscalibrated uncertainty quantification, biases towards superficial qualities such as answer positioning, style and tone, safety, numerical frequency and preferences, the type of underlying LLM being judged, and numerous other failure modes.

To mitigate these shortcomings, we developed \textscpop{Verdict}, a library for building compound LLM judge systems. \textscpop{Verdict} provides both the primitives (\textttpop{Units}) and execution framework for building such systems. Instead of a single LLM call to produce a judge result, \textscpop{Verdict} Judges combine multiple units of reasoning, verification, debate, and aggregation into a single judge system. When applied, these judge architectures leverage additional inference-time compute to yield impressive results on automatic evaluation of LLMs and LLM applications. 

\textscpop{Verdict}'s primary contributions are as follows:

\begin{enumerate}
    \item \textscpop{Verdict} provides a \textbf{unified interface} for a variety of prompting strategies, bias mitigation methods, and architectures grounded in frontier research. We support ideas from the disciplines of automated evaluation, scalable oversight, safety and content moderation, fact-checking, generative reward modeling, and more. 
    \item \textscpop{Verdict} introduces \textbf{powerful reasoning primitives and patterns} for automated evaluation, such as hierarchical reasoning verification and debate-aggregation.
    \item \textscpop{Verdict} is \textbf{fully composable}, allowing arbitrary reasoning patterns to be stacked into expressive and powerful architectures.
    \item \textscpop{Verdict} judges require \textit{minimal fitting} but achieve competitive performance on a wide variety of challenging automated evaluation tasks spanning safety moderation, factual and logical correctness, and hallucination detection.
\end{enumerate}

\section{Related Work}
% Prompted  Judges
\subsection{Prompted Judges}
Prompted out-of-the-box LLMs are the earliest and most straightforward approach to automated evaluation. This method involves providing an LLM with specific instructions or criteria to assess outputs \citep{bai2024benchmarking, lin2023llm, zheng2023judging}. The effectiveness of prompted judges has been demonstrated in both pairwise comparison and single output scoring scenarios. One key advantage of prompted judges is their flexibility. They can be easily adapted to evaluate different aspects of LLM outputs, such as linguistic quality, content accuracy, and task-specific metrics. However, the reliability of prompted judges is often inhibited by factors such as prompt design, formatting, and in-context example selection. In the same way that underlying LLMs suffer from sensitivity, brittleness, and hallucinations, so too do prompted LLM judges \citep{shankar2024validates, thakur2024judging, wang2023large}.

% Fine-Tuned Judges 
\subsection{Fine-Tuned Judges}
One alternative to prompted judges is fine-tuned custom judge models. These fine-tuned judges alleviate the biases and shortcomings of prompted judges, and sometimes are also cheaper and smaller than prompted judges. Indeed, it is possible to produce fine-tuned judge models that are only 7B parameters with performance on-par with leading closed-source models \citep{kim2023prometheus, kim2024prometheus, zhu2023judgelm}. Oftentimes, custom judges are fine-tuned to improve performance on specific tasks like fact-checking and hallucination detection, rather than general preference modeling and evaluation \citep{tang-etal-2024-minicheck, wang2024halu}.

The emerging study of generative verifiers and reward models has also provided new insights for LLM judging. Unlike traditional discriminative verifiers, generative reward models utilize next-token prediction for training, allowing it to tap into the benefits of generative LLMs. Research has demonstrated that generative reward models outperform discriminative verifiers, DPO verifiers, and LLM-as-a-Judge approaches \citep{ankner2024critique, zhang2024generative}. On algorithmic and math reasoning tasks, generative reward models show a 16-40\% improvement in the number of problems solved using the Best-of-N method. 

While these models can often times appear strong, they have been shown that to overfit their training data distribution \citep{huang2024empirical, sphynx2024}.

% These models are trained on synthetic data produced by more capable models, and support both direct judging as well as pairwise judging. 

% Early Compound Judges
\subsection{Early Compound Judges}
Recent approaches have leveraged multiple LLM calls to enhance the accuracy, consistency, and robustness of evaluations. One such method is LLM debate, where multiple instances argue from different perspectives, presenting arguments and counterarguments. A final judge LLM analyzes these debates to provide a comprehensive and balanced conclusion \citep{chan2023chateval, chu2024pre, li2023prd}. This technique has contemporaneously gained traction in the scalable oversight community, particularly for improving the accuracy of less-capable judges \citep{kenton2024scalable, michael2023debate}.

Ensemble judging, which independently queries multiple LLM judges and aggregates their assessments, has also become widely adopted. Aggregation strategies range from simple majority voting to weighted averaging based on model confidence, as well as more sophisticated fusion techniques that account for the strengths of different models \citep{liang2024abseval, verga2024replacing}.

These early compound judge systems have demonstrated significant improvements over single-LLM methods. Smaller, weaker models can often be combined to rival or even surpass the performance of larger models. \textscpop{Verdict} unifies and builds upon these insights to yield even more reliable, accurate, and powerful compound judge systems.

% \textscpop{Verdict} Core Concepts
\section{Overview: Core Concepts \& Orchestration}
At the heart of \textscpop{Verdict} are (1) primitives for judging (\textttpop{Units}), and (2) methods for linking, orchestrating, and executing systems of \textttpop{Units}.

% \section{\textscpop{Verdict} Primitives: Units}

\subsection{Basic Anatomy of a Unit}
A \textttpop{Unit} is the fundamental building block of a judge system. They are composed of the following:
\begin{itemize}
    \item \textbf{Prompt:} The instructions for how a \textttpop{Unit} should function.
    \item \textbf{Model:} The model that the \textttpop{Unit} calls to generate responses.
    \item \textbf{Scale:} The domain that generated values must be restricted to. For example, the \texttt{1-5} Likert scale, or the \textttpop{Yes/No} scale, or the \textttpop{Safe/Unsafe} scale, or any other ordinal or categorical scale.
    \item \textbf{Input Schema:} The values and types the \textttpop{Unit} can accept, either from a previous \textttpop{Unit} or from raw user-provided input.
    \item \textbf{Response Schema:} The values and types the \textttpop{Unit}'s model can generate.
    \item \textbf{Output Schema:} The values that are ultimately generated by the \textttpop{Unit}. This usually involves either postprocessing values produced by the Response Schema or referencing state from prior \textttpop{Units}. These values get passed to the subsequent \textttpop{Unit}.
\end{itemize}

\begin{figure}[t]
    \centering
    \includegraphics[width=0.9\linewidth]{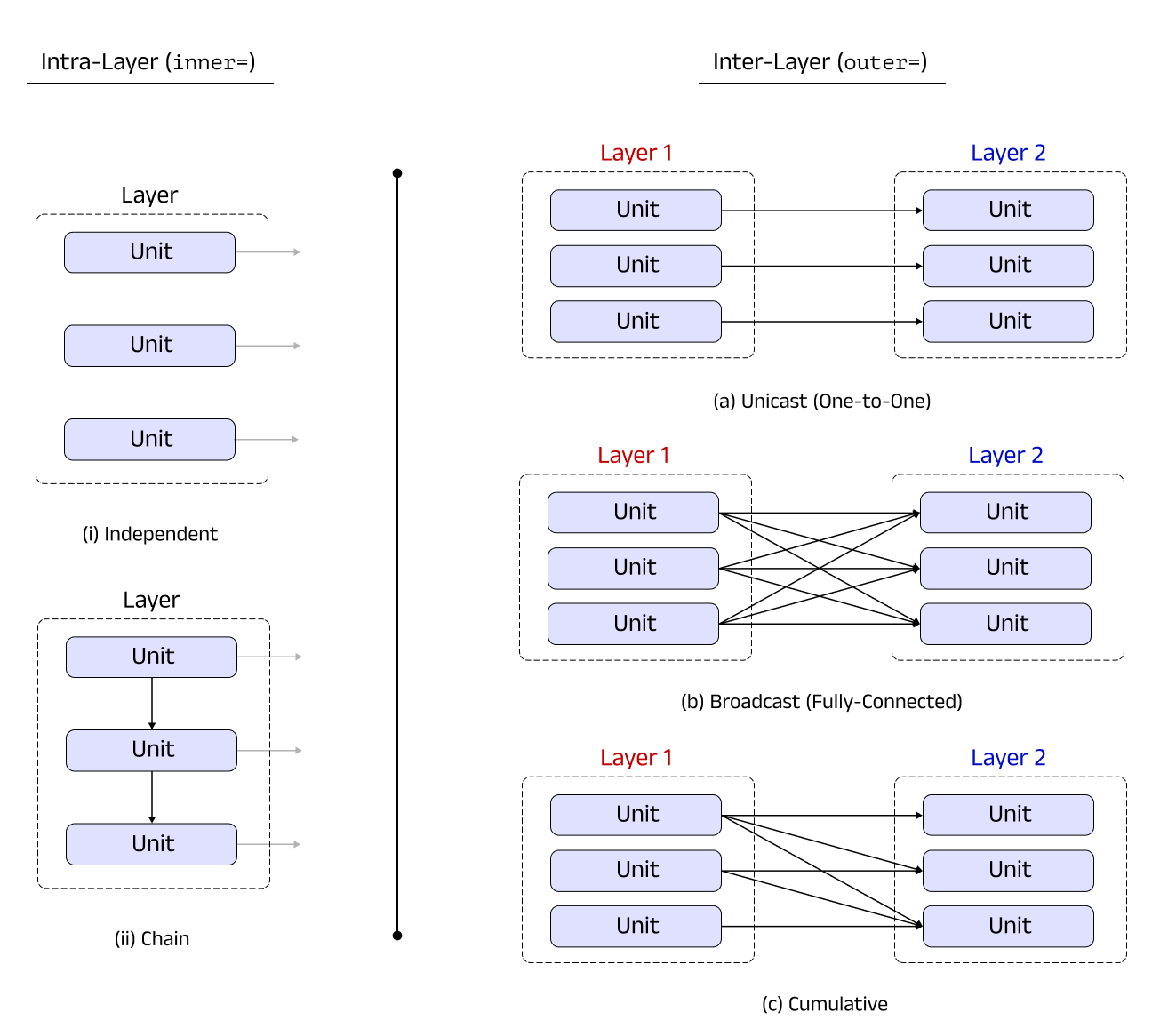}
    \caption{Primitives for linking \textttpop{Units} together within and between \textttpop{Layers}.}
    \label{fig:layer-connections}
\end{figure}

The structure of a \textttpop{Unit} solves two major pain points of standard LLM judges:
\begin{enumerate}
    \item The output structure of the judge is \emph{predictable} even for small language models, governed in particular by the Scale, Response Schema, and Output Schema.
    \item Format and function are specified separately. The Prompt implements the evaluation logic---the function---of the \textttpop{Unit}, while the Scale and Schemas manage the format of the \textttpop{Unit}.
\end{enumerate}

In the context of a \textscpop{Verdict} system, \textttpop{Units} that are chained together are also automatically type-checked. The Output Schema of a \textttpop{Unit} must match the Input Schema of its subsequent \textttpop{Unit}. This allows for information to flow through a compound judge system in a stable and predictable fashion.

\subsection{Structured Extraction}
While simple Scales like the \texttt{1-5} Likert Scale are straightforward to interpret and implement, they miss out on the rich information from a model's underlying probability distribution.

For more precise, calibrated, and powerful judging, \textttpop{Units} support accessing, manipulating, and aggregating the normalized log probabilities across their specified Scale.  One can \emph{extract} the log probability as a final score, or \emph{extract} a weighted sum (with potentially learned weights to further reduce bias) from the distribution \citep{liu2023g}. For simplicity, probabilities are not managed by default unless a \textttpop{Unit} invokes an Extractor.

\subsection{Execution Flow}
\textttpop{Units} on their own only re-implement existing LLM-judge methods. Stitching them together, however, unlocks the full power of \textscpop{Verdict} and increased inference-time compute. Combining \textttpop{Units} with the right architecture priors can yield impressive results across a variety of evaluation, judging, and reward modeling tasks.

\textscpop{Verdict} draws inspiration from deep learning libraries like PyTorch vis-\`a-vis managing \textttpop{Unit} groups and interactions. The most fundamental organizational principle is that of a \textttpop{Layer}, which is a list of \textttpop{Units}. This is analogous to how neural networks are computation graphs composed of submodules.

A standard \textttpop{Layer} propagates information through a judge system in a \emph{feedforward} fashion. By default, subsequent \textttpop{Layer's} \textttpop{Units} receive the output of previous \textttpop{Layer's} \textttpop{Units} in a one-to-one fashion, and \textttpop{Units} within a \textttpop{Layer} are fully independent of one another. However, it is possible to customize \textttpop{Unit} behavior both \emph{within} a \textttpop{Layer} (using keyword \textttpop{inner=}) and \emph{between} current and subsequent \textttpop{Layers} (using the \textit{outer} keyword). Figure \ref{fig:layer-connections} illustrates common ways in which \textttpop{Units} can be connected across \textttpop{Layers}. Verdict also implements a \textbf{MapUnit} to aggregate outputs from a previous \textttpop{Layer's} \textttpop{Units} (and optionally across several \textttpop{Layers}).

% Strong Results and Broad Applicability
\section{Results}

\begin{figure}
    \centering
    \includegraphics[width=\linewidth]{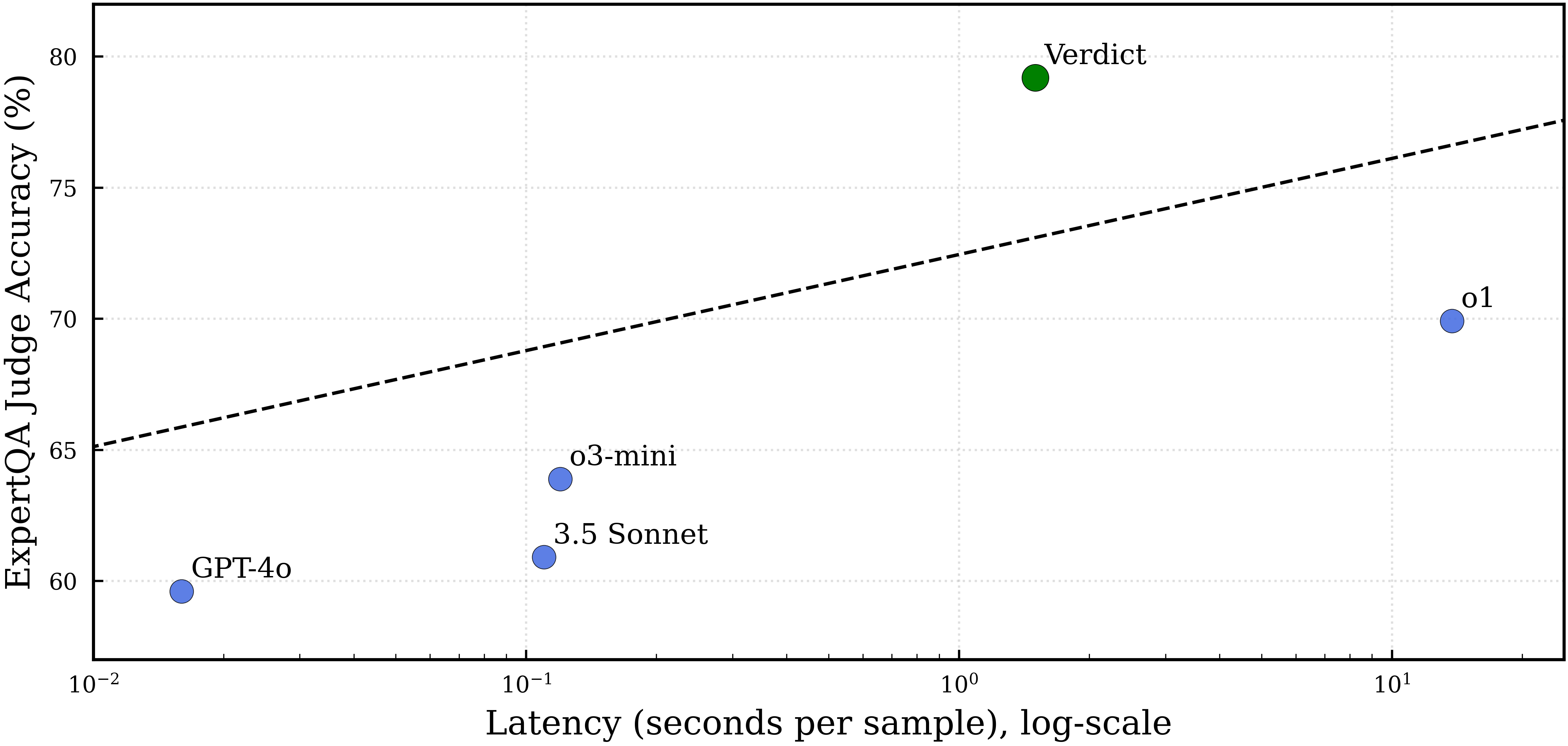}
    \caption{A simple \textsc{Verdict} judge surpasses reasoning models such as \texttt{o1} (+9.28\%) on the ExpertQA benchmark while operating at a fraction of the cost and latency. By explicitly programming the reasoning-trace \textit{structure} for each evaluation task, Verdict judges can be calibrated to a fixed computational budget. As shown above, Verdict pipelines can be tuned to approach or extend the Pareto frontier relative to their constituent prompted judges.}
    \label{fig:tradeoff}
\end{figure}

Across diverse evaluation settings ranging from content moderation, factual and logical validation, and hallucination detection, Verdict systems match the performance of specialized evaluator models and carefully prompted judges.

% Safety Moderation
\subsection{Safety Moderation}
\begin{wraptable}{c}{0.315\textwidth}
\vspace{-0.3in} % adjust vertical placement
\caption{Results on XSTest. The reported \textscpop{Verdict} system is a pipeline of \textttpop{CoTUnit} + \textttpop{JudgeUnits} + \textttpop{MeanPoolUnit}.}
\label{tab:xstest}
\vskip 0.05in
\begin{small}
\begin{tabular}{lc}
\toprule
Judge Model & Score \\
\midrule
\textscpop{Verdict} & \textbf{96.44\%} \\
o1 & 96.00\% \\
GPT-4o & 96.00\% \\
o1 Mini & 95.56\% \\
WildGuard & 95.11\% \\
o1 Preview & 94.89\% \\
GPT-4o Mini & 93.33\% \\
Llama-Guard-3-8B & 90.44\% \\
Claude 3.5 Sonnet & 87.33\% \\
Claude 3.5 Haiku & 84.44\% \\
Llama-Guard-2-8B & 83.56\% \\
Claude 3 Opus & 80.89\% \\
\bottomrule
\end{tabular}
\end{small}
\vspace{-0.15in}
\end{wraptable}

We first demonstrate the power of \textscpop{Verdict} on XSTest, a test suite containing challenging, borderline prompts that induce False Positive and False Negative safety classifications by LLMs. XSTest comprises (1) 250 safe prompts across ten prompt types that well-calibrated models should not refuse to comply with, and (2) 200 unsafe prompts as contrasts, which models should typically refuse for most applications.

XSTest includes diverse prompt types, such as homonyms and figurative language, to test various aspects of model behavior. Each prompt is crafted as a single English sentence in question format to simulate dialogue. XSTest helps identify exaggerated safety behaviors where models refuse to respond to safe prompts, evaluates model calibration by testing responses to both safe and unsafe prompts, and highlights systematic content moderation failure modes across leading LLMs. Table \ref{tab:xstest} shows the results of the \textscpop{Verdict} judge vis-à-vis other judges. Our judge is a system consisting of one \textttpop{CoTUnit}, two \textttpop{JudgeUnits}, and one \textttpop{MeanPoolUnit}.

% We show the prompts we used for non-\textscpop{Verdict} models in Appendix X.

% Factual and Logical Correctness
\subsection{Factual and Logical Validation}
\textscpop{Verdict} is also useful for evaluating the factual and logical correctness of LLMs. In Table \ref{tab:combined_results}, we showcase how \textscpop{Verdict} judges can achieve competitive performance on JudgeBench, a benchmark designed to assess LLM-based judges on complex responses spanning knowledge, reasoning, math, and coding. Each example in JudgeBench consists of a response pair with corresponding preference labels reflecting \textit{objective correctness}. This is unlike previous benchmarks that implicitly emphasize instruction-following and style preferences. JudgeBench is a challenging benchmark---many frontier models perform only marginally better than random guessing. However, a \textscpop{Verdict} architecture performs well, only second to the \texttt{o1} models.  

\begin{table*}[t]
\caption{\textit{Left:} Results on JudgeBench. Our \textscpop{Verdict} system consists of 4 rounds of debate and a \textttpop{JudgeUnit} that determines the final \textscpop{Verdict} based on these arguments. All \textttpop{Units} use GPT-4o. \textit{Right:} Results on the LLM-AggreFact ExpertQA split. Results are taken directly from the LLM-AggreFact leaderboard. Our \textscpop{Verdict} Judge is an ensemble of three instances of a \textttpop{JudgeUnit} with explanation followed by a self-verifier \textttpop{JudgeUnit}. The results of each instance are aggregated via a \textttpop{MaxPoolUnit}.}
\label{tab:combined_results}
\vskip 0.15in
\centering
\begin{small}

\begin{minipage}[t]{0.44\textwidth}
\centering
\begin{tabular}{lc}
\toprule
Judge & JudgeBench \\
\midrule
o1 & \textbf{75.43\%} \\
o1-mini & 65.71\% \\
Claude-3.5-Sonnet & 64.29\% \\
\textscpop{Verdict} (4o) & 63.55\% \\
Llama-3.1-405B & 56.86\% \\
GPT-4o & 56.57\% \\
Llama-3.1-70B & 52.29\% \\
GPT-4o-mini & 50.00\% \\
Gemini-1.5-pro & 47.14\% \\
Llama-3.1-8B & 40.86\% \\
Gemini-1.5-flash & 39.71\% \\
Claude-3-Haiku & 33.14\% \\
\bottomrule
\end{tabular}
\end{minipage}%
\hspace{0.08\textwidth}%
\begin{minipage}[t]{0.44\textwidth}
\centering
\begin{tabular}{lc}
\toprule
Judge & ExpertQA \\
\midrule
\textscpop{Verdict} (4o) & \textbf{79.17\%} \\
o1 & 69.91\% \\
\textscpop{Verdict} (4o-mini) & 67.72\% \\
GPT-4o & 64.67\% \\
o3-mini & 63.88\% \\
Claude-3.5 Sonnet & 60.90\% \\
Qwen2.5-72B & 60.10\% \\
QwQ-32B-Preview & 60.00\% \\
Bespoke-Minicheck-7B & 59.20\% \\
MiniCheck-Flan-T5-L & 59.00\% \\
Llama-3.1-405B & 58.50\% \\
Llama-3.3-70B & 58.30\% \\
\bottomrule
\end{tabular}
\end{minipage}

\end{small}
\vskip -0.1in
\end{table*}

% Hallucination Detection
\subsection{Hallucination Detection}
Table \ref{tab:combined_results} compares a \textscpop{Verdict} system---a triplet ensemble of \textttpop{JudgeUnit} models verified by another \textttpop{JudgeUnit} and aggregated with a \textttpop{MaxPoolUnit}---against other judges on the ExpertQA dataset. 

The ExpertQA dataset is designed to evaluate the factuality of AI-generated responses in domain-specific contexts. It consists of 2,177 expert-curated questions across medicine, law, history, and engineering. The dataset includes (1) expert-written questions that reflect real-world professional scenarios, (2) model-generated answers, evaluated by experts for factuality, attribution, and reliability, and (3) expert-revised answers, ensuring factual correctness and alignment with credible sources.

An answer that is deemed not factual by an expert is considered to be a hallucination. Notably, our \textscpop{Verdict} system using \texttt{GPT-4o} outperforms the SOTA judge (\texttt{GPT-4o}) by +14.5\%. The same \textscpop{Verdict} system, using a weaker backbone of \texttt{GPT-4o-mini}, still outperforms \texttt{GPT-4o} by +3.05\%.

% that of models specifically tuned for this particular task, with no manual tuning. On RAGTruth, we outperform all models excluding those specifically tuned for the task of hallucination detection.

\section{Applications}

Ostensibly, Verdict judges are used for offline evaluation, but practically speaking Verdict judges can be used anywhere to replace human feedback and verification. Naturally, they apply to at least the following scenarios:

\begin{enumerate}
    \item \textbf{Automated Evaluation} of AI Applications. Verdict judges enable tailored and automated evaluation of AI applications.
    \item Run-Time \textbf{Guardrails}. Verdict judges are guardrails that sit on top of AI applications running in production.
    \item \textbf{Test-Time Compute Scaling}. Verdict judges are verifiers that help rank, prune, and select candidates during test-time compute scaling.
    \item \textbf{Reward Modeling} \& Reinforcement Learning. Verdict judges provide signal in reinforcement learning --- particularly in settings where rewards are not easily verifiable.
\end{enumerate}

Given the recent innovation around scaling inference-time compute, reinforcement learning, and reasoning models, it is prudent to point out that \textscpop{Verdict} judges can indeed be leveraged as \textit{verifiers} in these settings. \textscpop{Verdict} is especially well-suited for verification, given that \textscpop{Verdict} judges are:
\begin{enumerate}
    \item More \textbf{general} than fine-tuned reward models. \textscpop{Verdict} judges readily apply across different tasks and domains, as seen by our experiments on safety moderation, checking for factual and logical correctness, and hallucination detection. 
    \item More \textbf{stable} and reliable compared to simple LLM judges. \textscpop{Verdict} judges beat out all simple LLM judges (and fine-tuned evaluators), barring the \texttt{o1} models on JudgeBench, on the three tasks presented here.
    \item Capable of generating \textbf{soft rewards}, unlike formal verifiers. This is critical for extending reasoning models beyond verifiable domains like mathematics and programming.
    \item Relatively \textbf{low-latency} and \textbf{cost-efficient} compared to similarly powerful judges, which is necessary for methods leveraging heavy inference-time compute.
\end{enumerate}
 
\section{Conclusion}
We introduce \textscpop{Verdict}: a modular, expressive, and flexible approach to automated evaluation of LLM outputs. By enabling the composition of diverse reasoning units---such as verification, debate, and aggregation---\textscpop{Verdict} enhances the robustness, interpretability, and accuracy of LLM judges.

\textscpop{Verdict} judges achieve competitive performance across a wide range of challenging evaluation tasks, including safety moderation, factual and logical verification, and hallucination detection.  Notably, \textscpop{Verdict}-based judges surpass both 1) models that are fine-tuned specifically for each evaluation task, as well as 2) orders-of-magnitude large foundation models when prompted. This highlights Verdict’s potential as an efficient and scalable alternative for AI evaluation.

Beyond the immediate demonstrated performance gains, Verdict serves as a unified framework for scaling judge-time compute. We hope Verdict will enable researchers and practitioners to develop reliable, accurate, and scalable AI evaluators to advance our field into the next era of major progress.

\section*{Acknowledgments}
We would like to thank Omar Khattab, Julian Michael, Jon Saad-Falcon, William Brown, Hamel Husain, Eugene Yan, and Shi Feng for their discussion and insights on scalable oversight protocols, automated evaluation, and generative reward models. We would also like to thank the gracious staff at Siena Bakehouse for fueling our work with delectable coffee and pastries.

\bibliography{colm2025_conference}
\bibliographystyle{colm2025_conference}

\appendix
\section{Appendix}
\subsection{Built-In Units}
Though \textttpop{Units} are meant to be maximally flexible and amenable to any practitioner or researcher use case, we implement the following base \textttpop{Units}.

\paragraph{Judge Units:}
\begin{itemize}
    \item \textbf{JudgeUnit:} Supports any discrete Scale, such as a Likert Scale, categorical Scale, or ordinal Scale. This is the standard LLM-as-a-judge.
    \item \textbf{PairwiseJudgeUnit:} Takes as input two strings and outputs a preferred choice between the two strings. This may also effectively be used as a categorical judge. 
\end{itemize}

\paragraph{Intermediate Units:}
\begin{itemize}
    \item \textbf{CoTUnit:} Generates a Chain of Thought in the usual sense.
    \item \textbf{ConversationalUnit:} Takes as input an existing conversation (list of messages) and generates the next turn of the conversation. Generally used in tandem with other \textttpop{ConversationalUnits}.
    \item \textbf{DebateUnit:} A subclass of \textttpop{ConversationalUnit} meant for debate (arguing opposing stances) between other \textttpop{DebateUnits}. Standard \textttpop{ConversationalUnits} are more neutral in their discourse.
\end{itemize}

\paragraph{Map/Aggregate Units:}
\begin{itemize}
    \item \textbf{MapUnit:} A generic way to aggregate outputs from a previous \textttpop{Layer's} \textttpop{Units} (and optionally across several \textttpop{Layers}).
    \item \textbf{MedianFilterUnit:} Takes the median of outputs from the previous layer of \textttpop{Units}.
    \item \textbf{MeanPoolUnit:} Takes the mean of outputs from the previous layer of \textttpop{Units}.
    \item \textbf{MaxPoolUnit:} Takes the max of outputs from the previous layer of \textttpop{Units}.
    \item \textbf{MeanVariancePoolUnit:} Takes the mean and variance of outputs from the previous layer of \textttpop{Units}.
\end{itemize}

% \begin{figure}[t]
%     \centering
%     \includegraphics[width=1\linewidth]{Units.png}
%     \caption{A few common \textttpop{Units} in \textscpop{Verdict}, segregated by functionality. \textttpop{JudgeUnits} produce a judge result (e.g. a float score or selected category); \textttpop{IntermediateUnits} produce reasoning as part of a broader \textscpop{Verdict} system; \textttpop{MapUnits} aggregate intermediate results.}
%     \label{fig:layer-connections}
% \end{figure}

\subsection{Examples of Compound Judge Systems}
To illustrate the expressiveness and ease of \textscpop{Verdict}, we use it to implement various examples from the LLM Judge literature with just a few lines of code. \footnote{These judge systems are also accessible via the \href{mailto:contact@haizelabs.com}{Haize Labs API}.}.

% Debate Example
\subsubsection{Debate}
In the field of scalable oversight, \textit{debates} are often used to evaluate LLM reasoning, resolve conflicting perspectives, and identify the most robust and well-supported answers by having multiple models critique and counter-argue each other's outputs \citep{khan2024debating, michael2023debate}.

The debate method involves multiple LLMs (typically two) arguing different sides of a question or task, with a judge LLM or human evaluating the debate to determine the most convincing argument. Typically, the process is as follows:

\begin{enumerate}
    \item A question or task is presented to the debater LLMs.
    \item Each debater LLM provides an initial answer and supporting arguments.
    \item In subsequent rounds, debaters respond to each other's arguments, providing additional evidence or pointing out flaws in the opponent's reasoning.
    \item After a set number of rounds (often 3--5), the judge reviews the entire debate transcript.
    \item The judge determines which debater presented the most convincing argument and selects a winner.
\end{enumerate}

\textscpop{Verdict} implements this relatively complex process with a \textttpop{Pipeline} comprising \textttpop{ConversationalUnits}, a \textttpop{MapUnit}, and a \textttpop{JudgeUnit}. \textttpop{ConversationalUnits}  sequentially on top of each to execute several rounds of debate. The \textttpop{MapUnit} aggregates information from the debate history into a string format, and finally the \textttpop{JudgeUnit} performs the binary classification to determine which \textttpop{ConversationalUnits} prevails.

\begin{figure}[h!]
    \centering
    \begin{mintedbox}{python}
debate_prompt = """
    You are participating in a debate as the {unit.role_name}.
    Here is the debate transcript thus far:
    {input.conversation}
"""

Pipeline("Debate") \
    >> Layer(
        [
            ConversationalUnit(role_name="Proponent").prompt(debate_prompt),
            ConversationalUnit(role_name="Opponent").prompt(debate_prompt)
        ],
        repeat=3,
        inner="chain",
        outer="last"
    ) \
    >> JudgeUnit(BooleanScale()).prompt("""
        Evaluate the following debate transcript:
        {previous.conversation}
    """)
    \end{mintedbox}
    \caption{A \textscpop{Verdict} \textttpop{Pipeline} for the Debate protocol. Leveraging \textttpop{ConversationalUnits}, \textttpop{MapUnits}, and \textttpop{JudgeUnits} makes for a quick and easy implementation.}
    \label{fig:debate}
\end{figure}

% Jury Example
\subsubsection{Jury: Ensemble of Judges}
Instead of using a single large model---like \texttt{o1}---as a judge, one can instead use a Panel of LLM evaluators (PoLL) composed of multiple smaller models \citep{verga2024replacing}. The PoLL approach of ensembling smaller, diverse language models offers a more cost-effective (seven times less expensive than GPT-4), less biased, and potentially more accurate method for evaluating LLM outputs.

% \begin{figure}[h!]
% \begin{mintedbox}{python}
% Pipeline("Jury") \
%     >> Layer(
%         [
%         JudgeUnit(DiscreteScale([1, 5])).prompt(qa_judge_prompt).via(model)
%             for model in [
%                 "gpt-4o-mini",
%                 "gpt-4o",
%                 "command-r",
%                 "command-r-plus",
%                 "claude-3-5-sonnet-20241022",
%             ]
%         ]
%     ) \
%     >> MeanPoolUnit()
%     \end{mintedbox}
%     \centering
%     % \includegraphics[width=1\linewidth]{examples/Jury.png}
%     \caption{A \textscpop{Verdict} \textttpop{Pipeline} implementation of Jury. Creating an ensemble of models is as simple as specifying their names in a list. Individual \textttpop{Unit} results are aggregated via the \textttpop{MeanPoolUnit}.}
%     \label{fig:jury}
% \end{figure}

\subsubsection{G-Eval}
% G-Eval \citep{liu2023g} is an automated evaluation framework for evaluating the quality of generated text in terms of accuracy, coherence, relevance, and factual alignment. With \textscpop{Verdict}, the core functionality of G-Eval can be expressed concisely, showcasing its ability to support advanced evaluation protocols with a simple, declarative structure.

G-Eval is an automated evaluation framework for natural language generation (NLG) tasks that uses LLMs with chain-of-thought (CoT) reasoning and a form-filling paradigm \citep{liu2023g}. It was developed to assess the quality of NLG outputs, such as text summaries and dialogue responses, with better human alignment than previous methods. Indeed, G-Eval with GPT-4 achieves a Spearman correlation of 0.514 with human evaluations on summarization tasks.

\textscpop{Verdict} can implement the G-Eval \textttpop{Pipeline} using a \textttpop{CoTUnit}, \textttpop{JudgeUnit}, and \textttpop{MeanVariancePoolUnit}. Critically, both here and for all pipelines implemented with \textscpop{Verdict}, the input and output schemas of each \textttpop{Unit} are validated and enforced throughout the \textttpop{Pipeline}. This allows the ML practitioner to focus on core judging logic without needing to worry about the brittleness of message passing between LLMs. 

\begin{figure}[h!]
    \centering
    \begin{mintedbox}{python}
scale = DiscreteScale((1, 5))
pipeline = Pipeline("GEval") \
    >> Layer(
        CoTUnit().prompt(f"""
            ### Generate evaluation steps for the following task:
            {TASK}
            ### Evaluation Criteria:
            {criteria_name} ({scale}) - {criteria_description}
            ### Evaluation Steps:
        """).via("gpt-4o", retries=3, temperature=0.6).pin() \
        # .pin(): run CoTUnit once and pass shared result to JudgeUnit across all samples

        >> JudgeUnit(scale).prompt(f"""
            {TASK}
            ### Evaluation Criteria:
            {criteria_name} ({scale}) - {criteria_description}
            ### Evaluation Steps:
            {{previous.thinking}}
            ### Source Text:
            {{source.source}}
            ### Summary:
            {{source.summary}}
            ### Evaluation Form (scores ONLY):
            - {criteria_name}:
        """).extract(WeightedSummedScoreExtractor()).via("gpt-4o-mini", retries=3, temperature=0.0)
    , repeat=5) \
    >> MeanVariancePoolUnit("score")
    \end{mintedbox}
    \caption{A \textscpop{Verdict} \textttpop{Pipeline} implementation of G-Eval using a \textttpop{CoTUnit}, \textttpop{JudgeUnit}, and \textttpop{MeanVariancePoolUnit}. Structure is enforced via the \textttpop{Scale} property, model parameters are easily managed on each \textttpop{Unit}, and the prompt can be flexibly defined inline.}
    \label{fig:g-eval}
\end{figure}

\end{document}